\documentclass[12pt]{article}
\usepackage{amsmath, setspace}
\usepackage{graphicx}
\usepackage{lineno}

 \makeatletter       
 \renewcommand{\@biblabel}[1]{#1.}
 \makeatother
\begin{document}
\noindent
{\LARGE\textbf{What does a convolutional neural network recognize in the moon?}}
\\
\\
{\large Daigo Shoji$^{1}$}\\
\\
1. Earth-Life Science Institute, Tokyo Institute of Technology, 2-12-1 Ookayama, Meguro-ku, Tokyo (shoji@elsi.jp)\\
\section*{Abstract}
Many people see a human face or animals in the pattern of the maria on the moon. Although the pattern corresponds to the actual variation in composition of the lunar surface, the culture and environment of each society influence the recognition of these objects (i.e., symbols)  as specific entities. In contrast, a convolutional neural network (CNN) recognizes objects from characteristic shapes in a training data set. Using CNN, this study evaluates the probabilities of the pattern of lunar maria categorized into the shape of a crab, a lion and a hare. If Mare Frigoris (a dark band on the moon) is included in the lunar image, the lion is recognized. However, in an image without Mare Frigoris, the hare has the highest probability of recognition. Thus, the recognition of objects similar to the lunar pattern depends on which part of the lunar maria is taken into account. In human recognition, before we find similarities between the lunar maria and objects such as animals, we may be persuaded in advance to see a particular image from our culture and environment and then adjust the lunar pattern to the shape of the imagined object.
\section{Introduction}
Because the periods of rotation and revolution are equal in the synchronous rotation of the moon, everyone on Earth sees almost the same lunar surface. The surface of the moon has color variation caused by the differences in composition. The large dark areas on the moon are called maria, which are composed of basalts erupted from the interior of the moon [1, 2]. 

\begin{figure}[htbp]
\centering
\includegraphics[width=10cm]{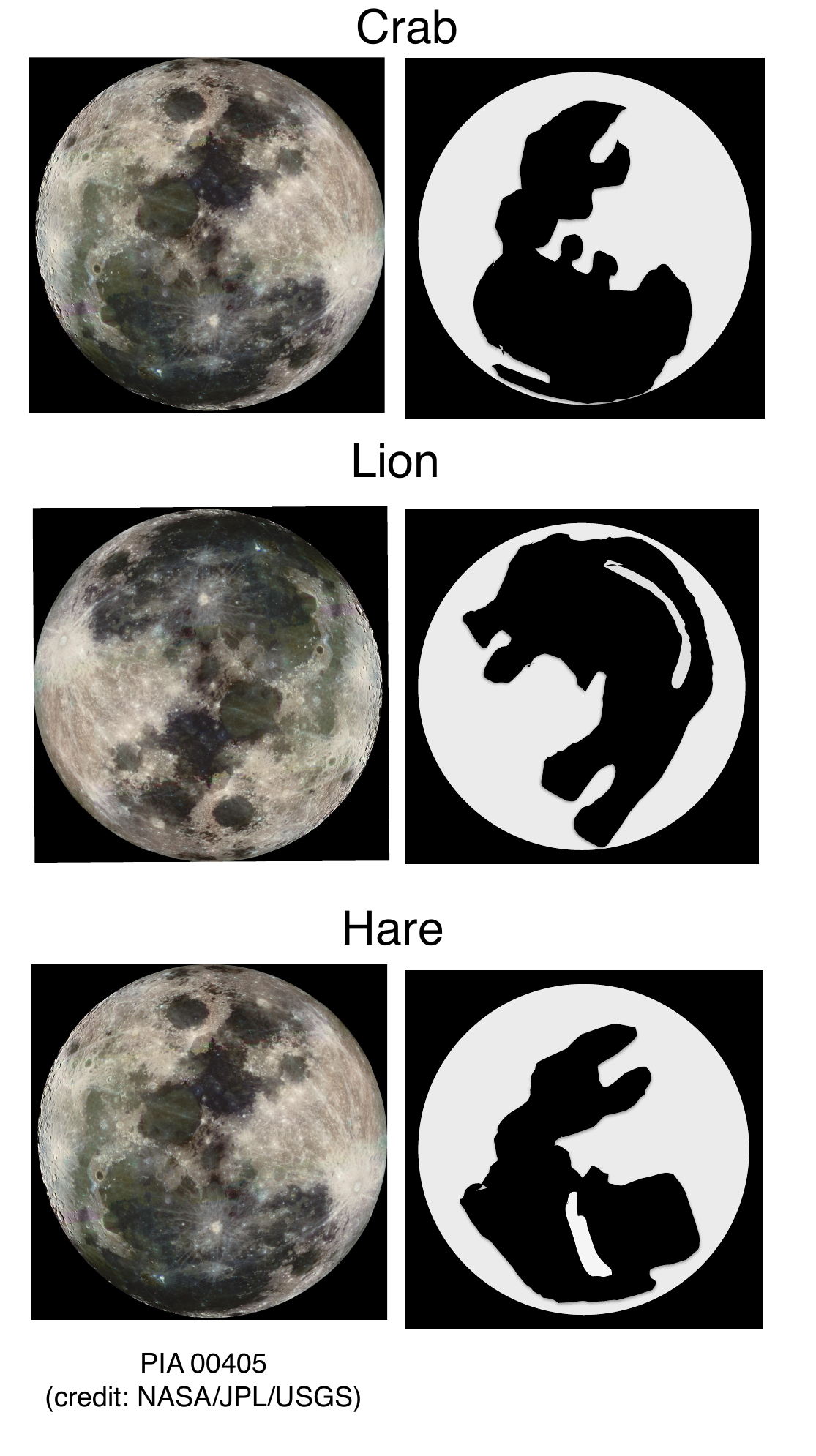}
\caption{Surface of the moon and the animals seen on the lunar surface. The illustrations are drawn based on the webpage of JAXA [3].}
\label{fig1}
\end{figure}

People recognize familiar objects from the pattern of the lunar maria, although the objects differ according to one's society and environment. In Europe, the lunar pattern is considered to be a human face [2]. On the other hand, the lunar pattern is considered to be a hare in Asian countries. In addition to the human face and the hare, a crab and a lion have also been reported as animals in the moon (Fig. \ref{fig1}) [3]. Although these objects in the moon are imagined due to the similarity between the lunar pattern and the shape of an animal, the culture and the environment in each society also influence the kind of object recognized in the moon. For example, in one of the old Buddhist tales, which are called "Jataka", a hare is exalted to the moon [4]. Thus, the hare in the moon believed in Asian countries is related to Buddhism. 

Recently, deep learning has achieved great success for recognizing and categorizing the objects in images. For example, deep learning of the CNN has been applied to the recognition of hand-written digits [5], human faces [6], and the detection of craters on Mars [7]. The CNN recognizes objects by learning the shape of training images. Thus, unlike human recognition, the CNN does not identify an object via culture and environment. 

In this work, shape recognition of the lunar pattern for the crab, lion and hare is performed by the CNN. First, the CNN learns the silhouette images, which consist of outlines of the three animals. By using silhouette images, even a simple CNN can categorize images with high accuracy and reduced computation time. Moreover, images of real animals are not required because the outline of the maria pattern is important. Evaluation using the real animal images is beyond the scope of this work. After training of the CNN and then testing the shapes of the lunar maria, the probabilities of the lunar pattern categorized to each animal are evaluated. It should be noted that the CNN does not learn the shape of the lunar pattern from the training of images. That is, the CNN recognizes the pattern of the lunar maria only by the shapes of the selected animals.

\section{Training of CNN}
\begin{figure}[htbp]
\centering
\includegraphics[width=17cm]{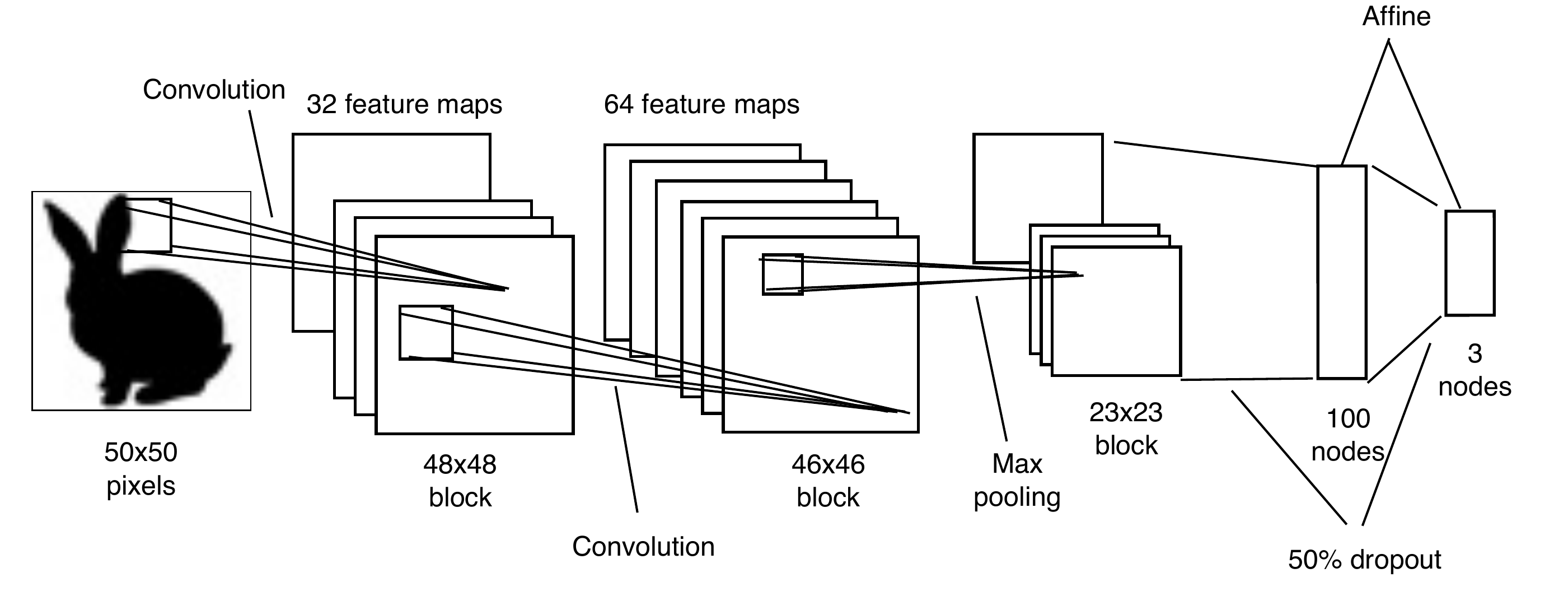}
\caption{Schematic view of the convolutional neural network. The network has two convolutional layers with 32 and 64 kernels, respectively. The size of a kernel is 3$\times$3 and the width of the stride is set at 1. After the max pooling layer and two affine layers, the probabilities are evaluated. Two dropout layers are included in front of the affine layers.}
\label{fig2}
\end{figure}

The structure of the CNN is shown in Fig. \ref{fig2}. The structure is composed of two convolutional layers and two affine layers. Training and testing of images are conducted with the Keras library [8]. The kernel size of the two convolutional layers is set at 3$\times$3, and the pooling layer decreases the size of the featured maps by 50\%. The optimizer is Adam, and loss functions and probabilities are calculated by cross-entropy and soft-max functions, respectively [8].

\begin{figure}[htbp]
\centering
\includegraphics[width=17cm]{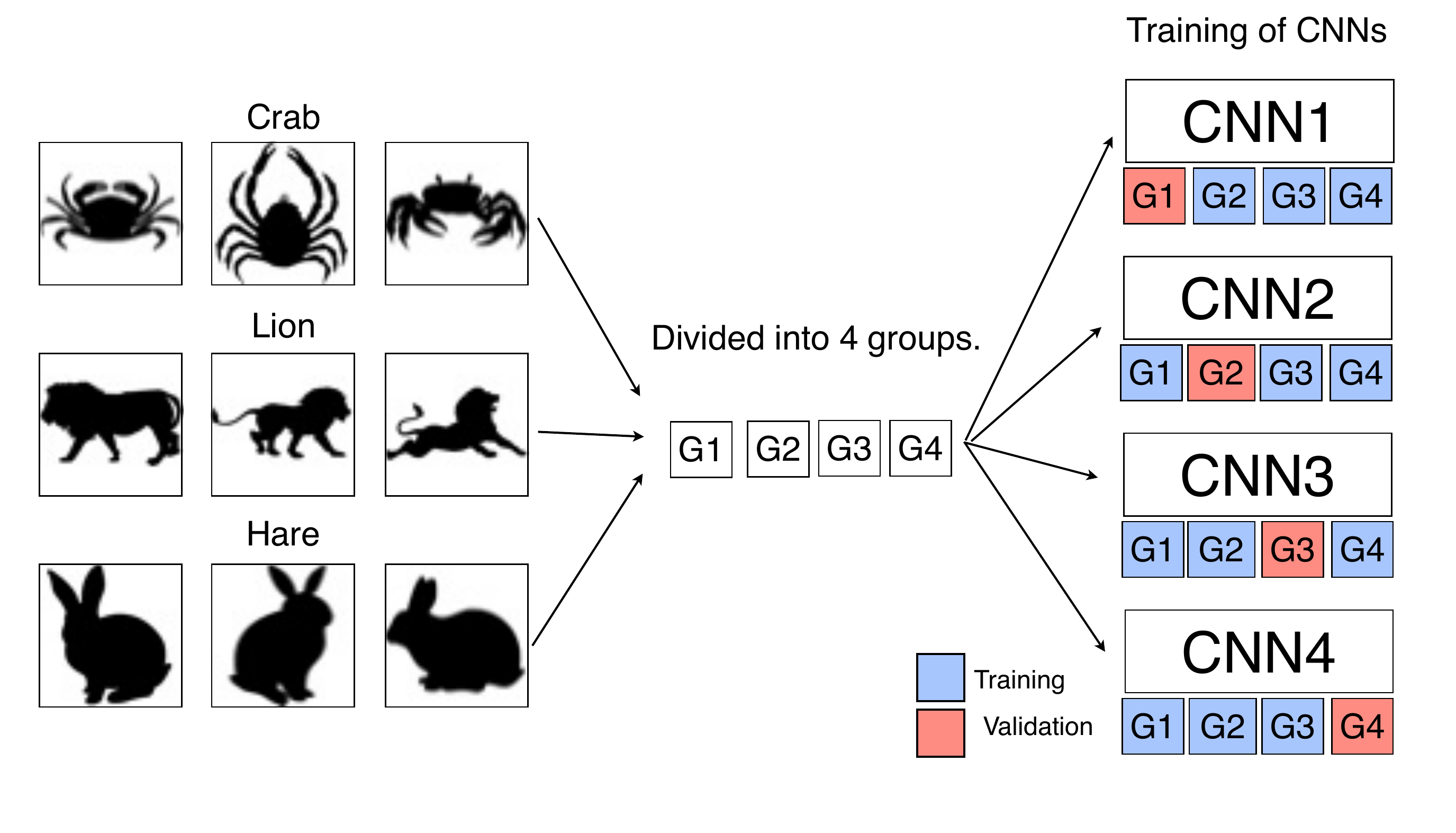}
\caption{Examples of silhouette images of the animals and the schematic view of the CNN training. Training images are randomly divided into four groups (G1, G2, G3 and G4). Each CNN uses three groups as the training data sets and one group as the validation data set. The group of the validation data set is different for each CNN. Weights of CNNs are conserved when the loss function of the validation data set is the minimum value.} 
\label{fig3}
\end{figure}

The numbers of collected silhouette images of the crab, lion and hare were 100, 100 and 105, respectively (I bought silhouette images from Deposit Photos (https://jp.depositphotos.com/), and downloaded free images from Silhouette AC (https://www.silhouette-ac.com/index.html), Illust AC (https://www.ac-illust.com), Clipart Library (http://clipart-library.com), Silhouette Design (http://kage-design.com/)). Example images are shown in Fig. \ref{fig3}. Of these 305 images, 240 images (80 images in each class) were used for training of the CNNs and 65 images were used as the test data set. The size of each image was 50$\times$50 pixels and the intensity of the images was normalized between 0 and 1 by dividing each intensity by 255. 

\begin{figure}[htbp]
\centering
\includegraphics[width=15cm]{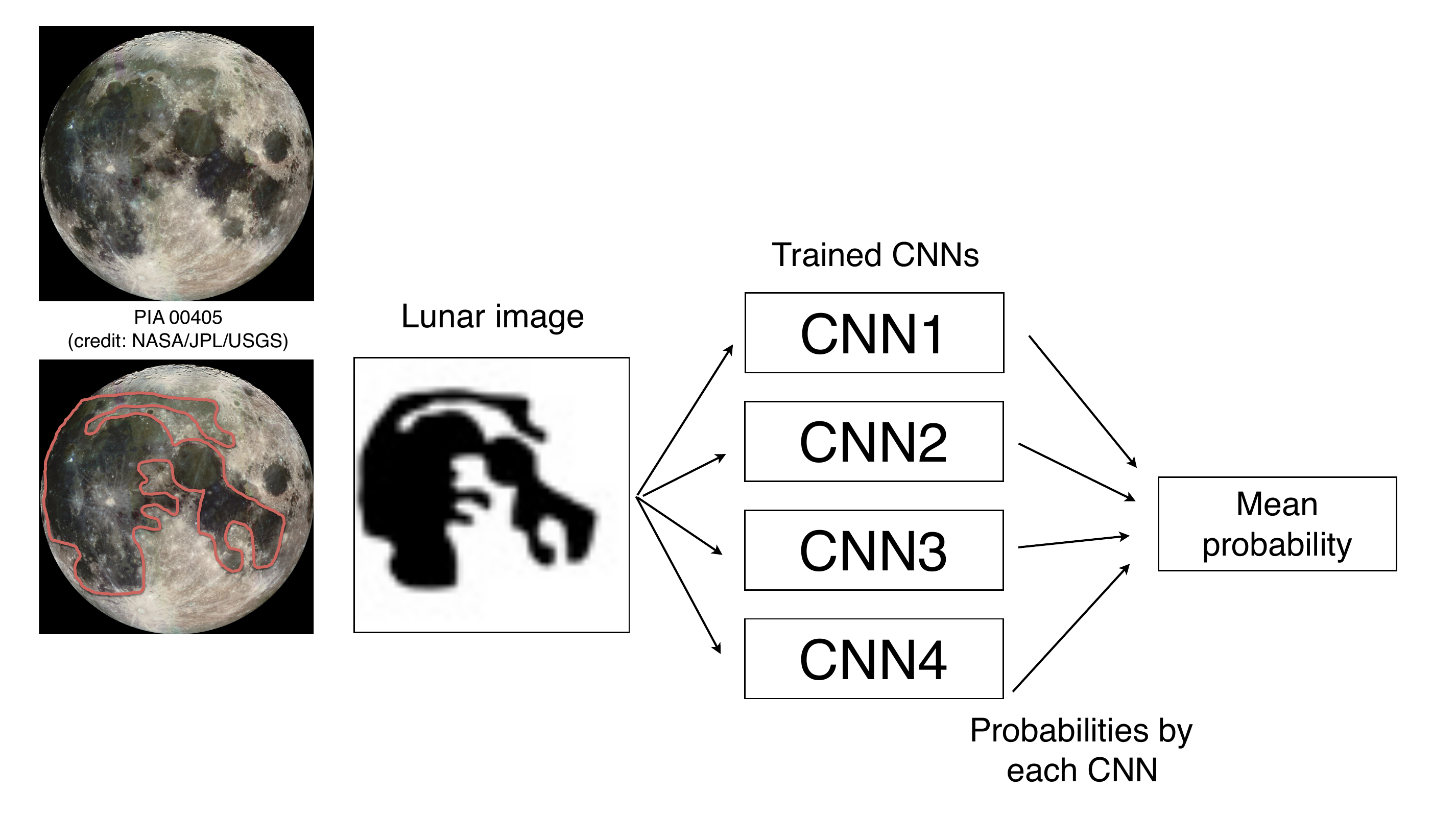}
\caption{Schematic view of the evaluation of probabilities of the lunar pattern. For the lunar images, the outline of the lunar maria is considered. Each CNN with the trained weights calculates the probabilities of the lunar image categorized as one of the three animals. The mean value of each probability is regarded as the final probability. }
\label{fig4}
\end{figure}

To increase the accuracy even with the small number of images, k-fold cross validation was used. A training data set with 240 images was randomly divided into four groups (60 images per group), and four CNNs with the same structure were trained by using three groups (180 images) as training data sets and one group (60 images) as a validation data set (Fig. \ref{fig3}). In each training of CNN, the group of validation data was changed. Each training was conducted up to 40 epochs, and the number of steps at one epoch to update the weights of the CNNs was set at 20 [8]. Within 40 epochs, weights resulting in the minimum loss function of the validation data set were conserved.  Because the number of training images was not sufficient, the batch size was set at 180 and the training images were increased by rotating between -180$^{\circ}$ and +180$^{\circ}$ and by shifting up to 10\% of the width and height of the images [8]. After trainings of the CNNs, the accuracy of the test data set was calculated from the mean probability by using the conserved weights of the four CNNs (Fig. \ref{fig4}).

An image of the lunar maria is shown in Fig. \ref{fig4}. The lunar maria are represented in black color and other effects such as the round shape of the moon itself and the color variations in the maria are removed because the purpose of this work was to evaluate the probability that the CNN can recognize and categorize the lunar maria pattern into the three animals. 

\section{Results}
For up to 40 epochs, each CNN recognized the validation images with 92\%-98\% accuracy. By using the trained weights and the mean probabilities of the four networks, the CNNs classified 61 of 65 test images correctly, and thus the accuracy was approximately 93.8\%. 

\begin{figure}[htbp]
\centering
\includegraphics[width=10cm]{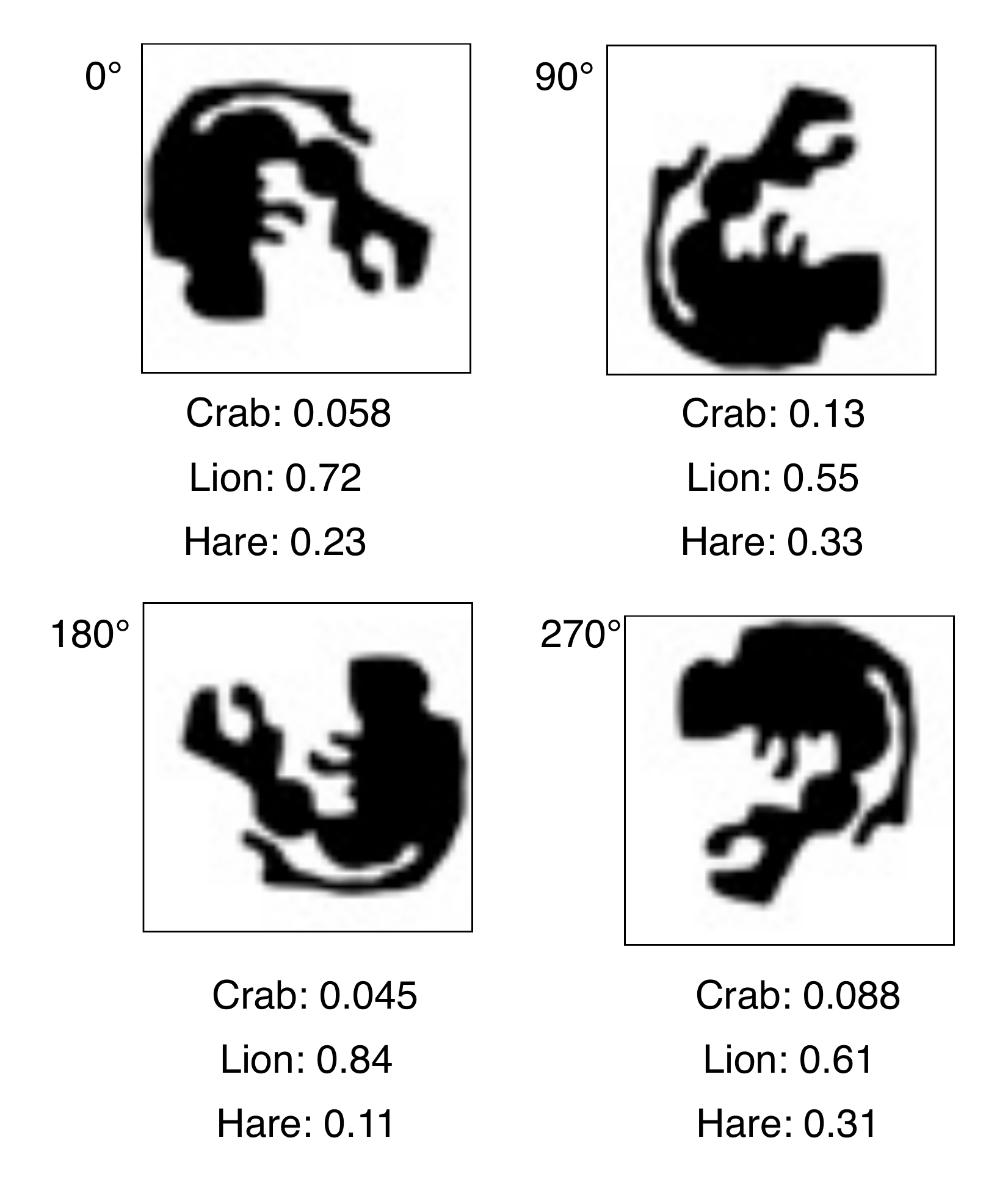}
\caption{Probabilities of the lunar pattern with Mare Frigoris for the three animals. Images are tested from four angles. Probabilities are rounded to two significant digits.}
\label{fig5}
\end{figure}

The probabilities of a lunar image categorized into the three animals are shown in Fig. \ref{fig5}. Each image of the lunar maria was tested four times by rotating the image four times by 90 degrees. At every rotation angle, the probability of categorization of the lion was the highest, and the CNN recognized that the lunar pattern was not the shape of the crab. 

\begin{figure}[htbp]
\centering
\includegraphics[width=10cm]{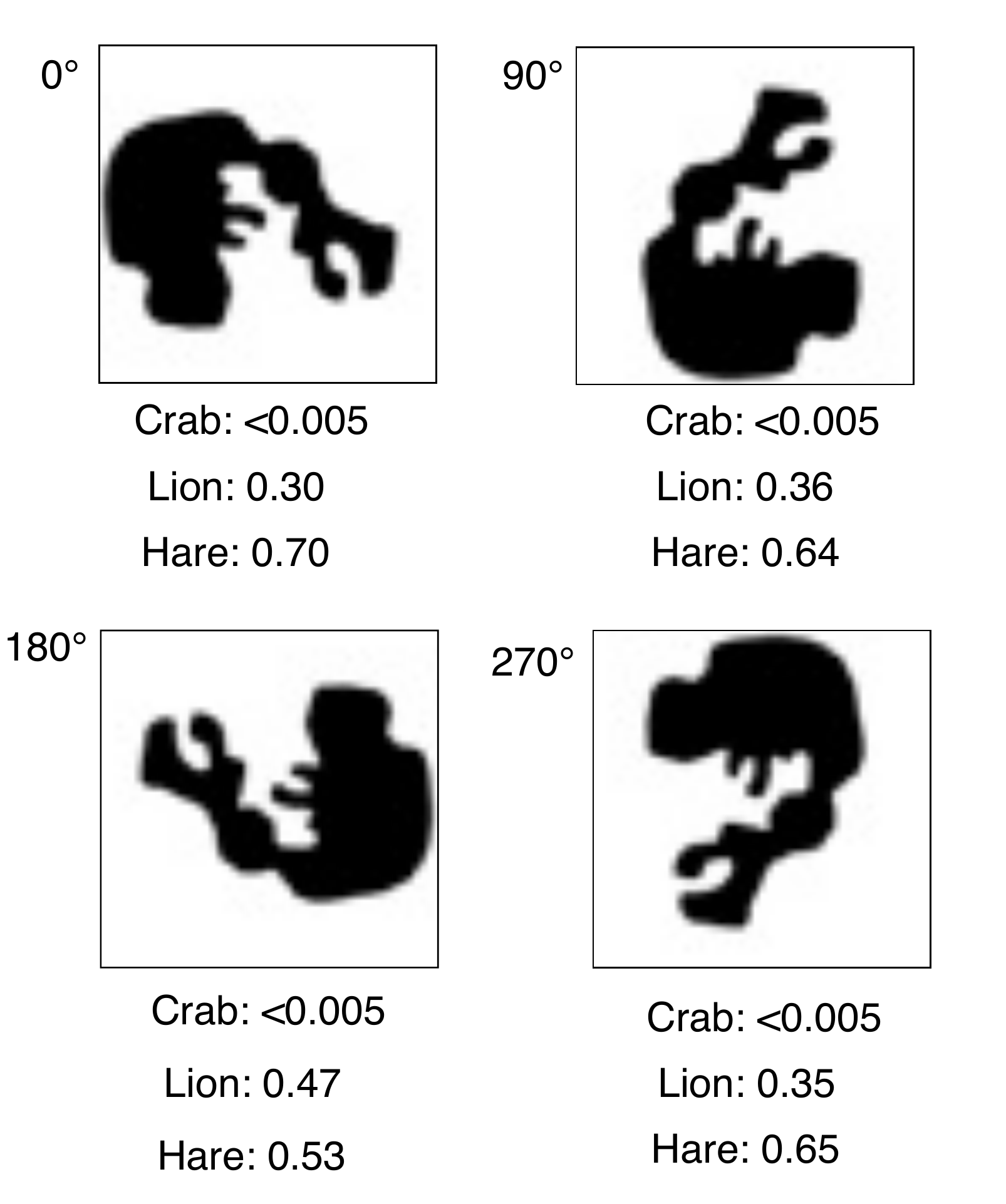}
\caption{Probabilities of the lunar pattern without Mare Frigoris for the three animals. Images are tested from four angles. Probabilities are rounded to two significant digits.}
\label{fig6}
\end{figure}

In the case of the lion, the dark band on the moon corresponding to Mare Frigoris [1] is regarded as the tail of the lion (Fig. \ref{fig1}). The shapes of the crab and the hare do not contain the band of Mare Frigoris. Fig. \ref{fig6} shows the probabilities of the lunar image without Mare Frigoris. The probabilities to the hare become highest when the dark band of Mare Frigoris is not included in the lunar pattern. Most silhouette images of the lion contain the lion's tail (Fig. \ref{fig3}). Thus, when the band is removed, the CNN concludes that the lunar image is the shape of the hare rather than that of the lion. 

In addition to the increased probabilities of the hare, the probabilities of categorizing the crab decreased drastically to almost zero (Fig. \ref{fig6}). The images of the crab used in this work contain legs as the primary shape of the crab (Fig. \ref{fig3}). In the case of the lunar pattern with Mare Frigoris, the CNN might recognize a crab by its claw  legs, and so the probability decreases for the lunar image without Mare Frigoris. Thus, the reason why the probabilities of categorizing the crab is the lowest in both lunar images is that the CNN learned the claw legs as the characteristic shape of the crab during training and the lunar images did not have sufficient areas related to the these legs.  

\section{Conclusion}
The probability of recognizing and categorizing the lunar pattern into three animals (crab, lion, hare) depends on which part of the lunar maria is included in the image. Of the three animals, the pattern of the lunar maria is recognized as the shape of the lion if the dark band of Mare Frigoris is included in the lunar pattern. On the other hand, the probability of recognizing the hare becomes the highest for the pattern without the band, and the probability of recognizing the crab is almost zero because the CNN recognize the area of Mare Frigoris as part of the crab's leg. 

The CNN classifies the lunar pattern from the characteristic shapes of the training images. Thus, if the lunar pattern is different from the shapes of the training images, the probability of recognition becomes low, as in the case of the crab in this work. This process is different from recognition by humans. Although people as well as CNN recognize the similarity of shapes, we can also relate different objects by their character (this may be known as "hierophany", shown in the work of Mircea Eliade). This character cannot be derived from the shape itself. For example, the moon,  the crab and the hare have been symbols  of immortality, rebirth and fertility  in many areas of the world [9], and thus the moon pattern is recognized as such animals via these symbolic characters. These characters are induced from the behaviors of the animals and the moon rather than from their shapes. Myths also influence the relationship between different objects, as shown by the "Jataka" tales in Buddhism [4]. In areas where lions do not live, it is difficult for people to connect a lion to the moon. Thus, the environment of a society is also important. 

If the culture and the environment of each society influence our recognition of the lunar pattern, we may decide what we see in the moon in advance, and then we may find the similarities of shapes by adjusting the lunar pattern (e.g., by removing the area of Mare Frigoris for the hare's shape or seeing the crab's shape without legs), which is opposite to the recognition mechanism of the CNN. However, we cannot connect objects that are completely different from the shape of the lunar pattern. Thus, feedback may exist between the similarity of shape and the influence of culture for the recognition of symbols (culture induces people to find similarity of some objects, and if they can find rough similarity, additional cultures such as myth are generated in the society, which strengthens the similarity unconsciously in people's mind). 

\section*{Acknowledgements}
This work was supported by a JSPS Research Fellowship. I bought silhouette images from deposit photos (https://jp.depositphotos.com/), and downloaded free images from Silhouette AC (https://www.silhouette-ac.com/index.html), Illust AC (https://www.ac-illust.com), Clipart Library (http://clipart-library.com), Silhouette Desighn (http://kage-design.com/).

\end{document}